\documentclass[journal]{IEEEtran}
\usepackage{cite}
\usepackage{amsmath,amssymb,amsfonts}
\usepackage{graphicx}
\usepackage{textcomp}
\usepackage{xcolor}
\def\BibTeX{{\rm B\kern-.05em{\sc i\kern-.025em b}\kern-.08em
    T\kern-.1667em\lower.7ex\hbox{E}\kern-.125emX}}
\usepackage{enumitem}
\usepackage[ruled,vlined]{algorithm2e}
\usepackage{booktabs}
\usepackage{makecell}
\usepackage{multirow}
\usepackage{hyperref}

\begin{document}
\title{Physically-Induced Atmospheric Adversarial Perturbations: Enhancing Transferability and Robustness in Remote Sensing Image Classification}

\author{Weiwei Zhuang, Wangze Xie, Qi Zhang, Xia Du,~\IEEEmembership{Member,~IEEE}, Zihan Lin, Zheng Lin, Hanlin Cai, Jizhe Zhou, Zihan Fang, Chi-man Pun,~\IEEEmembership{Senior Member, ~IEEE}, Wei Ni,~\IEEEmembership{Fellow,~IEEE}, and Jun Luo,~\IEEEmembership{Fellow,~IEEE}
\thanks{Weiwei Zhuang, Wangze Xie, and Xia Du are with the School of Computer and Information Engineering, Xiamen University of Technology, Xiamen, 361000, China (e-mail: zhuangweiwei@xmut.edu.cn; xiewangze@stu.xmut.edu.cn; duxia@xmut.edu.cn).}
\thanks{Qi Zhang is with the Faculty of Data Science, City University of Macau, Macau SAR, China (e-mail: qizhang@cityu.edu.mo).}
\thanks{Zihan Lin is with the Dundee International Institute, Central South University, Changsha, China (e-mail: zihan-lin@csu.edu.cn).}
\thanks{Zheng Lin is with the Interdisciplinary Centre for Security, Reliability and
Trust (SnT), University of Luxembourg, L-1855 Luxembourg, Luxembourg
(e-mail: zhenglin@ieee.org).}
\thanks{Hanlin Cai is with the Department of Engineering, University of Cambridge, Cambridge, UK (e-mail: hc663@cam.ac.uk).}
\thanks{Zihan Fang is with Hong Kong JC STEM Lab of Smart City and Department of Computer Science, City University of Hong Kong, Kowloon, Hong Kong SAR, China (e-mail: zihanfang3-c@my.cityu.edu.hk).}
\thanks{Jizhe Zhou and Chi-man Pun are with the Department of Computer and Information Science, Faculty of Science and Technology, University of Macau, Macau, 999078, China (e-mails: yb87409@um.edu.mo, cmpun@umac.mo)}
\thanks{Wei Ni is with the School of Engineering, Edith Cowan University, Perth, WA 6027, Australia (email: wei.ni@ieee.org).}
\thanks{Jun Luo is with the College of Computing
and Data Science, Nanyang Technological University, Singapore (e-mail: junluo@ntu.edu.sg).}
\thanks{Co-corresponding author: Wangze Xie (xiewangze@stu.xmut.edu.cn); Xia Du (duxia@xmut.edu.cn); Zheng Lin (linzheng@eee.hku.hk).}}



\maketitle

\begin{abstract}
Adversarial attacks pose a severe threat to the reliability of deep learning models in remote sensing (RS) image classification. Most existing methods rely on direct pixel-wise perturbations, failing to exploit the inherent atmospheric characteristics of RS imagery or survive real-world image degradations. In this paper, we propose FogFool, a physically plausible adversarial framework that generates fog-based perturbations by iteratively optimizing atmospheric patterns based on Perlin noise. By modeling fog formations with natural, irregular structures, FogFool generates adversarial examples that are not only visually consistent with authentic RS scenes but also deceptive. By leveraging the spatial coherence and mid-to-low-frequency nature of atmospheric phenomena, FogFool embeds adversarial information into structural features shared across diverse architectures. Extensive experiments on two benchmark RS datasets demonstrate that FogFool achieves superior performance: not only does it exceed in white-box settings but also exhibits exceptional black-box transferability (reaching 83.74\% TASR) and robustness against common preprocessing-based defenses such as JPEG compression and filtering. Detailed analyses, including confusion matrices and Class Activation Map (CAM) visualizations, reveal that our atmospheric-driven perturbations induce a universal shift in model attention. These results indicate that FogFool represents a practical, stealthy, and highly persistent threat to RS classification systems, providing a robust benchmark for evaluating model reliability in complex environments.
\end{abstract}


\section{Introduction}
With the rapid development of satellite and airborne sensors, vast amounts of remote sensing imagery with high spatial, spectral, and temporal resolutions are continuously collected \cite{zhang2022progress,lin2024fedsn,yuan2024satsense,peng2024sums,lin2025leo,zhao2024leo,yuan2023graph}, providing critical information for a wide range of applications, such as land cover mapping \cite{slomp2025improving}, urban planning \cite{yu2022combined}, environmental monitoring \cite{shu2025deep}, and disaster detection \cite{qin2021landslide}. Effectively extracting semantic information from these data is a key challenge in remote sensing image analysis. Among downstream tasks, remote sensing image classification is fundamental, aiming to assign semantic labels to image pixels or regions based on visual and spectral information \cite{paheding2024advancing}. Recent advances in deep learning, particularly convolutional neural networks (CNNs)~\cite{lin2025hsplitlora,fang2025dynamic,sun2025rrto,fang2026nsc,lin2024efficient}, have substantially improved classification performance through automatic feature learning \cite{cheng2020remote}. While these methods achieve impressive results on benchmark datasets \cite{thapa2023deep}, their vulnerability to adversarial perturbations raises concerns about robustness and reliability in real-world applications \cite{xu2023ai,mei2024comprehensive}.

Adversarial attacks have recently attracted increasing attention in the remote sensing community as deep learning techniques become widely adopted in this field. Czaja et al. \cite{czaja2018adversarial} conducted one of the earliest empirical studies on adversarial examples in satellite imagery, highlighting the unique challenges of remote sensing scenarios, including viewing geometry, atmospheric effects, and temporal variability. Subsequent studies demonstrated that remote sensing image classification models are highly susceptible to adversarial perturbations under both white-box and black-box settings, posing serious security risks in practical applications. However, most existing attack methods rely on direct pixel-wise perturbations and fail to exploit the inherent characteristics of remote sensing imagery, such as ubiquitous atmospheric phenomena. This reliance on high-frequency digital noise often results in limited cross-model transferability and low resistance to image-processing defenses, making them less effective in practical RS pipelines.

Although several studies have explored weather-based attacks, notable limitations persist. For example, Tang et al. \cite{tang2023natural} simulated snow, fog, shadows, and sun flares to attack optical aerial object detectors; however, their perturbations mainly imitate weather-related color patterns and fail to capture the irregular and natural structures of real clouds and shadows, resulting in visually artificial shapes, e.g., circles or triangles. Sun et al. \cite{sun2024defense} proposed an adversarial cloud attack for remote sensing salient object detection by constraining images with exposure matrices and additive perturbations, but the generated cloud patterns still exhibit noticeable unnatural noise. Moreover, both methods are restricted to object detection tasks. For remote sensing image classification, Ma et al. \cite{ma2025cloud} introduced a cloud-based adversarial example generation method that uses a Perlin Gradient Generator Network optimized via differential evolution to enable query-efficient black-box attacks. Nevertheless, this approach does not support targeted attacks, lacks controllability over cloud density, and requires additional network training for parameter generation.

\begin{figure}[t]
    \centering
    \includegraphics[width=\linewidth]{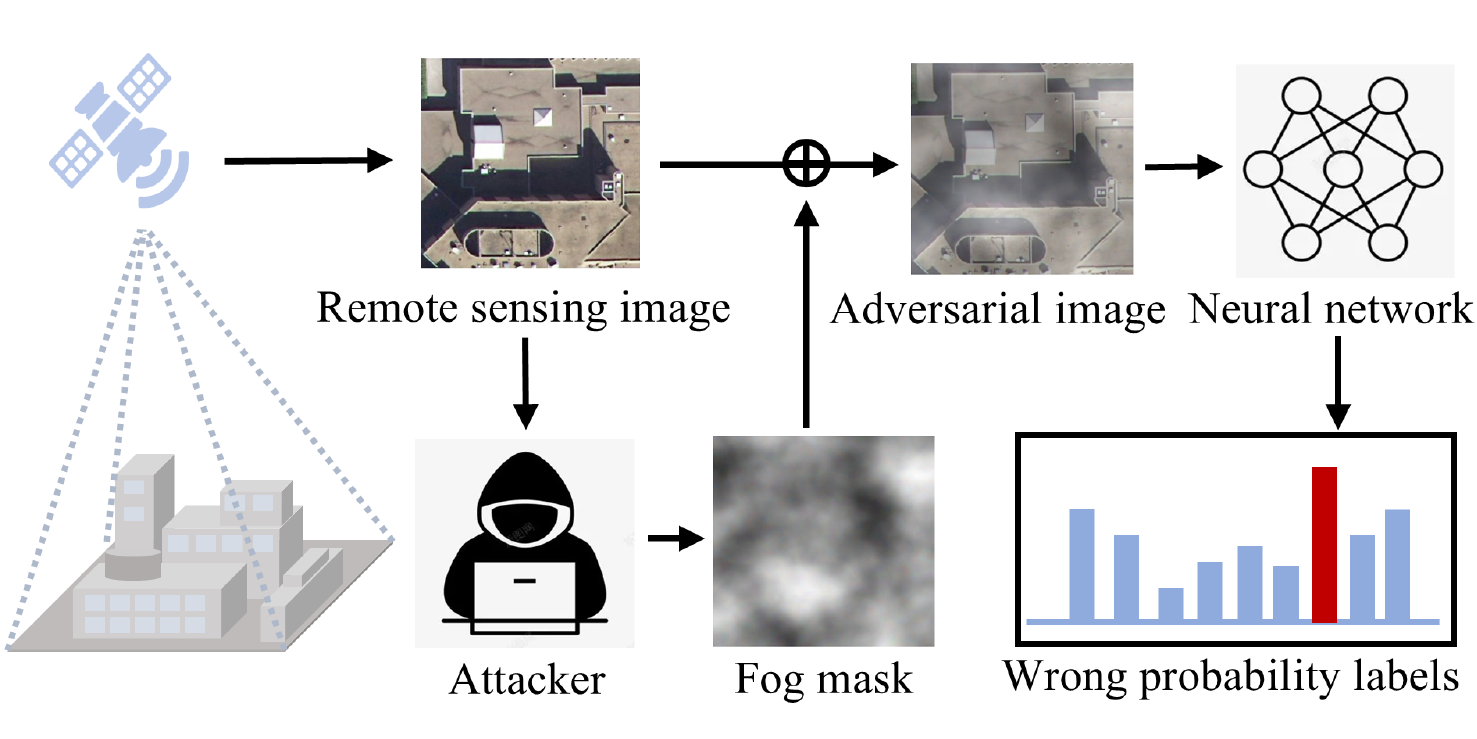}
    \caption{An overview of the proposed fog-based adversarial attack scenario in remote sensing image classification. An attacker generates a physically plausible fog mask using Perlin noise, which is blended with the original remote sensing image to form an adversarial example. When fed into a deep neural network, this adversarial example causes the model to produce incorrect classification results, posing a security threat to real-world remote sensing systems.}
    \label{fig:scenario}
\end{figure}

Motivated by these limitations, this paper investigates adversarial attacks that exploit atmospheric phenomena in remote sensing imagery. As illustrated in Fig. \ref{fig:scenario}, the adversarial threat to remote sensing classification systems can be realized through physically plausible atmospheric perturbations. In this scenario, an attacker optimizes a natural-looking fog mask based on procedural Perlin noise, which is then blended with the original image to create an adversarial example. By modeling fog formations with natural, irregular structures, the generated adversarial examples remain visually consistent with authentic remote sensing images. By shifting the perturbation from isolated pixels to spatially coherent atmospheric structures, FogFool targets the mid-to-low frequency features shared across diverse deep learning architectures, hence significantly enhancing black-box transferability.

Compared to existing methods, the main contributions of this paper are summarized as follows:
\begin{enumerate}
    \item We propose \textit{FogFool}, a fog-based adversarial attack framework for remote sensing image classification that generates targeted and untargeted adversarial examples by optimizing fog patterns using Perlin noise, thereby producing structured, visually natural perturbations.
    \item We design a controllable fog-density parameterization that enables fine-grained control over perturbation strength and facilitates the analysis of the trade-off between visual realism and attack effectiveness.
    \item Extensive experiments on two benchmark remote sensing datasets validate that FogFool yields superior attack performance with 97.67\% / 99.93\% targeted success rates on UCM/NWPU and 83.74\% black-box transfer rate on NWPU, while exhibiting stronger robustness against JPEG compression and filtering defenses by embedding adversarial information into global atmospheric distributions.
\end{enumerate}

The remainder of this paper is organized as follows: Section~\ref{sec:related_work} reviews related work on adversarial attacks. Section~\ref{sec:method} details the proposed FogFool framework. Section~\ref{sec:experiments} presents experimental settings and results. Section~\ref{sec:conclusion} concludes this paper.

\section{Related Work}
\label{sec:related_work}

\subsection{Adversarial Attacks on Deep Neural Networks}
Adversarial examples were first revealed by \cite{szegedy2013intriguing}, which demonstrated that imperceptible perturbations can cause deep neural networks to make incorrect predictions with high confidence. Since then, extensive studies have explored the vulnerability of deep models to adversarial examples. Existing attack methods are generally categorized into optimization-based approaches, such as L-BFGS \cite{szegedy2013intriguing} and C\&W attacks \cite{carlini2017towards}, and gradient-based approaches, including FGSM \cite{goodfellow2014explaining}, its iterative variants such as BIM \cite{kurakin2018adversarial} and PGD \cite{madry2017towards}, as well as momentum-based methods \cite{dong2018boosting}. Beyond digital-domain attacks, recent studies have explored physically realizable and visually natural adversarial perturbations, e.g., adversarial stickers \cite{wei2022adversarial}, shadow-based attacks \cite{zhong2022shadows}, weather-induced disturbances, including rain and snow \cite{marchisio2022fakeweather}, and adversarial camouflage that blends perturbations into natural textures and styles \cite{duan2020adversarial}.

\subsection{Adversarial Attacks in Remote Sensing}
Under the white-box assumption, several studies have systematically investigated the vulnerability of remote sensing image classification models. Chen et al. \cite{chen2019adversarial} first evaluated classical gradient-based attacks on CNN-based remote sensing image recognition models, revealing their sensitivity to adversarial perturbations. They further observed that adversarial misclassifications tend to be selective, with predictions biased toward semantically similar classes. Subsequent large-scale empirical analyses confirmed the prevalence of adversarial examples across different network architectures and datasets \cite{chen2021empirical}, and demonstrated their strong transferability across models \cite{xu2020assessing}.

\begin{figure*}[!t]
    \centering
    \includegraphics[width=1\linewidth]{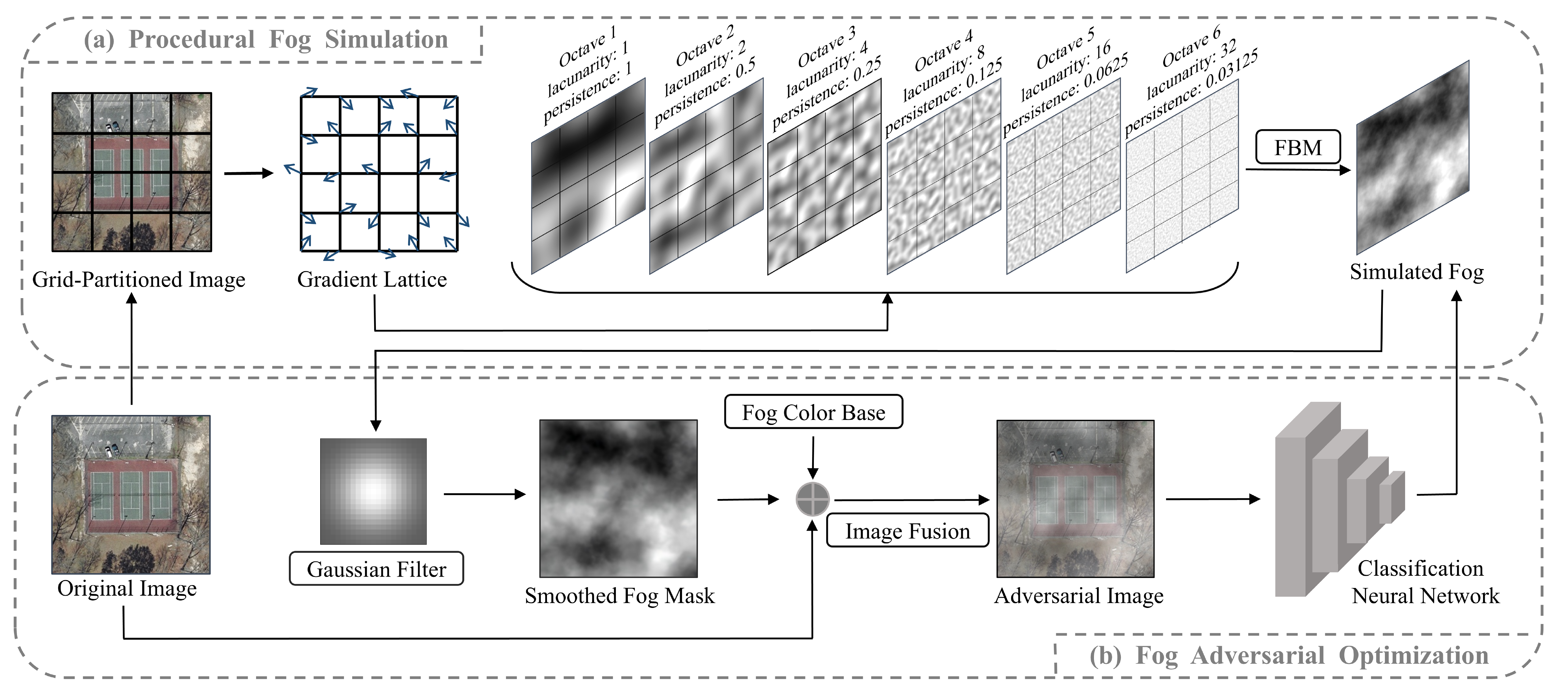} 
    \caption{Overview of the proposed method for fog adversarial example generation with (a) Procedural Fog Simulation Module, (b) Gradient-Guided Fog Adversarial Optimization Module}
    \label{fig:framework}
\end{figure*}

In contrast to white-box attacks, black-box attacks are more practical in real-world scenarios, where the internal structure and parameters of the target model are inaccessible. A key strategy for black-box adversarial attacks is to enhance the transferability of adversarial examples across different models \cite{papernot2016transferability}. Xu and Ghamisi \cite{xu2022universal} pioneered the study of universal adversarial examples in remote sensing, proposing black-box attack methods that exploit shared vulnerabilities among models and establishing a benchmark for evaluation. Building upon this line of research, Bai et al. \cite{bai2022targeted} introduced targeted variants of universal adversarial examples, enabling more fine-grained and controllable attacks. More recently, Wang et al. \cite{wang2025ppca} proposed a black-box attack framework based on feature approximation that improves perturbation precision and cross-model transferability. These studies have collectively demonstrated that black-box adversarial attacks pose a serious and realistic threat to remote sensing image classification models.

\section{Atmospherically Inspired Adversarial Perturbation Design}
\label{sec:method}

\subsection{Overview of the Proposed Method}
As illustrated in Fig.~\ref{fig:framework}, we propose a fog-based adversarial attack framework, namely FogFool, that exploits the procedural characteristics of Perlin noise to simulate realistic atmospheric fog and iteratively optimize it for effective adversarial perturbation. The overall framework consists of two major components: a Procedural Fog Simulation Module and a Gradient-Guided Fog Adversarial Optimization Module.

Given a clean remote sensing image as input, the proposed FogFool method first constructs a procedural fog representation using Perlin noise. Specifically, the image is partitioned into a grid, forming a grid-partitioned image, with random gradient vectors assigned to each grid vertex, yielding a gradient lattice. Based on this lattice, multiple Perlin noise maps are generated for different combinations of persistence and lacunarity, yielding six noise maps corresponding to different octaves. These multi-scale noise maps are then fused using fractional brownian motion (FBM) to produce a single fog-intensity map, referred to as the Simulated Fog\footnote{The fusion of multi-octave Perlin noise via FBM is designed to mimic the multi-scale fractal characteristics of real atmospheric fog, which is more physically plausible than single-scale noise-based fog simulation and avoids artificial grid artifacts common in conventional perturbation methods.}. This process enables modeling of spatially coherent, multi-scale fog patterns that resemble real atmospheric phenomena.

In the adversarial optimization stage, the simulated fog is further smoothed via Gaussian filtering to obtain a Smoothed Fog Mask, which is then blended with a predefined fog color base and the original remote sensing image via image fusion, generating a fog-based adversarial example. The adversarial image is then fed into the target classification network, and the loss gradient with respect to the fog parameters is computed. Guided by the model gradients, the simulated fog is iteratively refined to maximize the attack effectiveness while maintaining visual plausibility. After multiple iterations, the optimized fog perturbation yields adversarial examples that can successfully mislead the classification model.

\subsection{Problem Description}
Let $f(\cdot)$ denote a trained image classification model that maps an input remote sensing image 
$\mathbf{x} \in \mathbb{R}^{H \times W \times C}$ to a probability distribution over $K$ classes, \textit{i.e.},
$f(\mathbf{x}) = \mathbf{p} \in [0,1]^K,$
where $\sum_{k=1}^{K} p_k = 1$. The predicted label is given by $\hat{y} = \arg\max_{k} f_k(\mathbf{x}).$

Given a clean image $\mathbf{x}$ with the ground-truth label $y$, the objective of the adversarial attack is to generate an adversarial example $\mathbf{x}^{adv}$ such that the model prediction is altered, while the applied perturbation 
remains visually imperceptible or physically plausible.

\subsubsection{Untargeted Adversarial Attack}
In an untargeted attack, the goal is to cause misclassification without specifying a target class, which can be formulated as
\begin{equation}
    \hat{y}^{adv} = \arg\max_{k} f_k(\mathbf{x}^{adv}),       ~~s.t.~\hat{y}^{adv} \neq y.
\end{equation}

This objective can be equivalently expressed by maximizing the classification loss:
\begin{equation}
\mathbf{x}^{adv} = \arg\max_{\mathbf{x}' \in \mathcal{S}} 
\mathcal{L}\big(f(\mathbf{x}'), y\big),
\label{eq:untargeted_attack}
\end{equation}
where $\mathcal{L}(\cdot)$ denotes the loss function (e.g., cross-entropy), and $\mathcal{S}$ represents the feasible perturbation space.

\subsubsection{Targeted Adversarial Attack}
In a targeted attack, the adversary aims to force the model to predict a specific target label $y_t \neq y$, \textit{i.e.}, $\hat{y}^{adv} = y_t$. Accordingly, the adversarial example is obtained by minimizing the loss with respect to the target class:
\begin{equation}
\mathbf{x}^{adv} = \arg\min_{\mathbf{x}' \in \mathcal{S}} 
\mathcal{L}\big(f(\mathbf{x}'), y_t\big).
\label{eq:targeted_attack}
\end{equation}

In this work, the feasible set $\mathcal{S}$ is constrained to structured and physically motivated perturbations 
corresponding to atmospheric fog effects. The detailed construction and optimization of such fog-based perturbations 
are introduced in the following sections.

\subsection{Perlin Noise}
\begin{figure}
    \centering
    \includegraphics[width=1\linewidth]{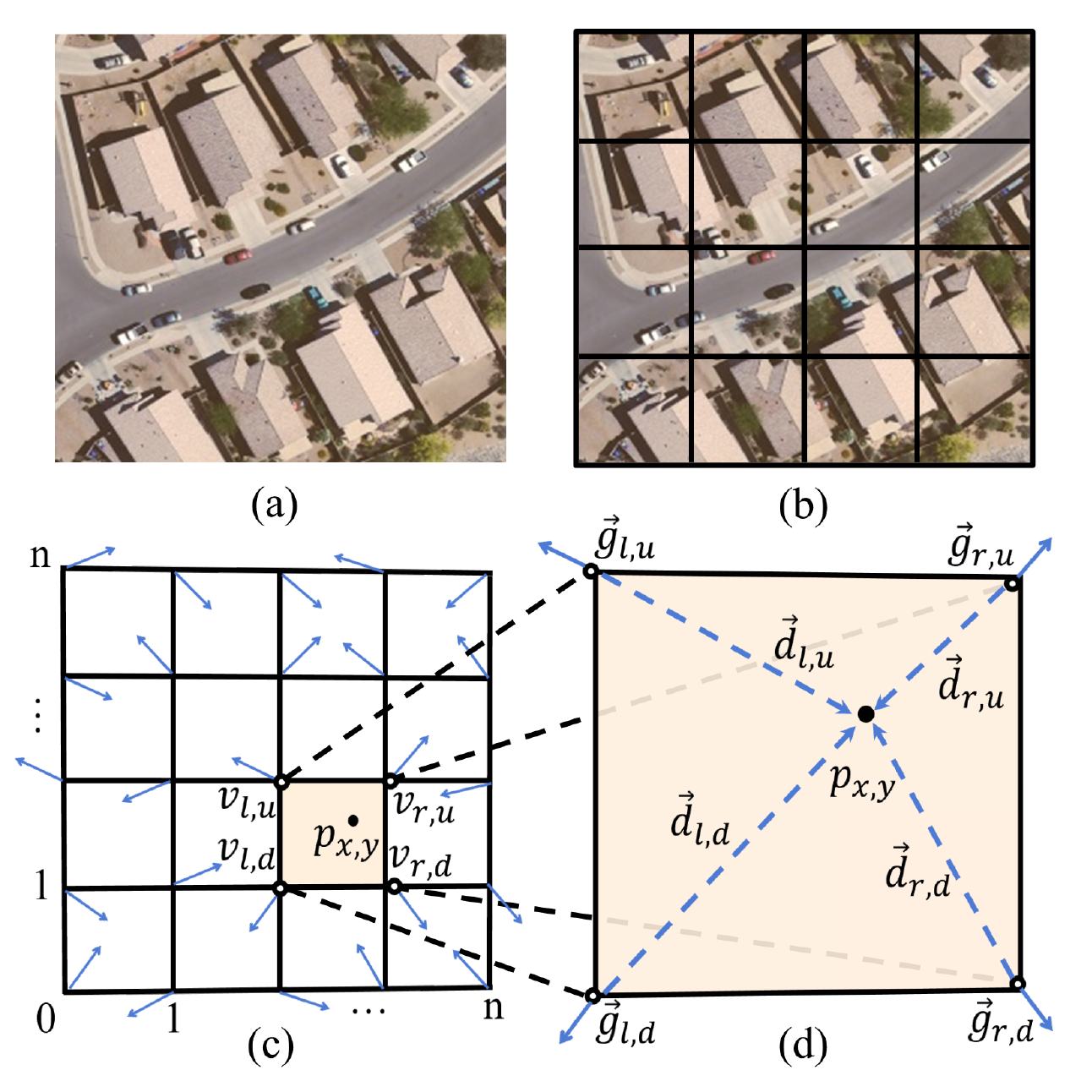}
    \caption{Process of 2-D Perlin noise generation. (a) The original remote sensing image. (b) The image is partitioned into an $n \times n$ grid. (c) Gradient vector assignment. (d) Gradient vectors $\vec{g}$ and displacement vectors $\vec{d}.$}
    \label{fig:perlin_noise}
\end{figure}
To simulate realistic atmospheric fog patterns, we adopt Perlin noise as the fundamental procedural modeling technique. Originally introduced by Perlin~\cite{perlin1985image} and later refined in~\cite{perlin2002improving}, Perlin noise is a gradient-based noise function that generates spatially coherent and continuous random fields. Due to its ability to produce natural-looking textures with controllable smoothness and scale, Perlin noise has been widely used in computer graphics to model natural phenomena, such as clouds, smoke, and fog.

In this work, Perlin noise is employed to construct a physically plausible fog-intensity distribution for remote-sensing imagery. The noise generation process consists of three main steps: grid partition and gradient lattice construction, gradient influence computation at arbitrary query points, and smooth interpolation for noise evaluation. To capture the inherent multi-scale characteristics of atmospheric fog, multiple Perlin noise maps are aggregated using FBM. The detailed formulation of each step is described as follows.

\begin{enumerate}[label=\textbf{Step \arabic*}:, leftmargin=*]

\item The process begins with the original remote sensing imagery, as displayed in Fig.~\ref{fig:perlin_noise}(a). The image domain is partitioned into a square lattice with dimensions $n \times n$, overlaying the grid structure shown in Fig.~\ref{fig:perlin_noise}(b). Subsequently, a random 2-D gradient vector is allocated to each node (intersection) of the grid, establishing the noise basis visualized in the schematic grid in Fig.~\ref{fig:perlin_noise}(c).

\item Let $P = \{p_{x,y} | 0 \le x, y < n\}$ be the set of query points. For an arbitrary point $p_{x,y}$ (located within the grid in Fig.~\ref{fig:perlin_noise}(c)), the surrounding unit square is delimited by four vertices $V = \{v_{i,j}| i = l, r; j = d, u\}$. The coordinate bounds are calculated as $l = \lfloor x \rfloor$ and $d = \lfloor y \rfloor$, with $r = l+1$ and $u = d+1$, where $\lfloor \cdot \rfloor$ denotes the floor operation. As detailed in the magnified view in Fig.~\ref{fig:perlin_noise}(d), let $\vec{g}_{i,j}$ denote the random gradient vectors at the vertices. The displacement vector, representing the offset from a grid corner $v_{i,j}$ to the target point $p_{x,y}$, is expressed as:
\begin{equation}
    \vec{d}_{i,j} = [x - i, y - j].
    \label{eq:desplacement_vector}
\end{equation}
Finally, the scalar influence $s_{i,j}$ (gradient ramp value) is derived for each vertex via the dot product of the displacement and gradient vectors:
\begin{equation}
    s_{i,j} = \vec{d}_{i,j} \cdot \vec{g}_{i,j}.
    \label{eq:gradient_ramp_value}
\end{equation}

\item After obtaining the gradient ramp values $s_{i,j}$ at the four surrounding vertices, the Perlin noise value at an arbitrary query point $p_{x,y}$ is computed through a smooth interpolation process.
To ensure continuity and suppress grid artifacts, a quintic smoothing function, commonly referred to as the fade function, was proposed in \cite{perlin2002improving} and is applied to the normalized local coordinates:
\begin{equation}
    \text{fade}(t) = 6t^{5} - 15t^{4} + 10t^{3}.
    \label{eq:fade_function}
\end{equation}
Let~$\Delta x = x - l$ and $ \Delta y = y - d$ denote the relative offsets of the query point within the unit grid cell. The fade function is then evaluated at $\Delta x$ and $\Delta y$ to produce smooth interpolation weights. Subsequently, linear interpolation is utilized to blend the gradient values smoothly, where the function $\text{lerp}(a, b, t)$ serves as a fundamental linear interpolation operator. Formally, it is defined as:
\begin{equation}
    \text{lerp}(a, b, t) = a + t(b - a).  
    \label{eq:interpolation_function}
\end{equation}
First, the gradient ramp values along the horizontal direction are interpolated at the lower and upper edges of the grid cell:
\begin{equation}
    \begin{aligned}
        l_{d} = \text{lerp}\big(s_{l,d}, s_{r,d}, \text{fade}(\Delta x)\big),\\
        l_{u} = \text{lerp}\big(s_{l,u}, s_{r,u}, \text{fade}(\Delta x)\big).
    \end{aligned}
    \label{eq:lerp_horizontal}
\end{equation}
Subsequently, a second interpolation is carried out along the vertical direction to obtain the final Perlin noise value at $p_{x,y}$:
\begin{equation}
f(x, y) = \text{lerp}\big(l_{d}, l_{u}, \text{fade}(\Delta y)\big).
\label{eq:lerp_vertical}
\end{equation}
Through this two-stage interpolation process, the resulting noise field is continuous and smooth across grid boundaries, forming the basic Perlin noise map used in the subsequent multi-octave fog modeling.

\end{enumerate}

While a single-layer Perlin noise produces spatially smooth but relatively simple patterns, realistic atmospheric fog exhibits multi-scale structures with both coarse and fine variations. To capture such characteristics, multiple Perlin noise maps at different spatial frequencies are combined using FBM.

Formally, the FBM noise field is defined as a weighted summation of Perlin noise functions across multiple octaves:
\begin{equation}
\mathrm{FBM}(x, y) = \sum_{k=0}^{K-1} \alpha_k \cdot f\big(\beta_k x, \beta_k y\big),
\label{eq:fbm_function}
\end{equation}
where $f(\cdot)$ denotes the base Perlin noise function defined in Eq.~(\ref{eq:lerp_vertical}), $K$ represents the total number of octaves, $\alpha_k$ is the amplitude scaling factor (persistence), and $\beta_k$ is the frequency scaling factor (lacunarity) at the $k$-th octave.

In this work, we adopt a commonly used exponential scaling strategy for FBM to balance visual realism and computational efficiency, following the standard Perlin noise configuration \cite{perlin1985image}. The number of octaves is set to $K = 6$ to sufficiently model the multi-scale structure of natural fog without incurring excessive computational cost. The amplitude factor $\alpha_k = 2^{-k}$ progressively weakens high-frequency fine details, consistent with the natural attenuation of atmospheric features, while the frequency factor $\beta_k = 2^{k}$ doubles the spatial frequency per octave to generate spatially coherent, realistic fog patterns.

By aggregating Perlin noise maps at increasing frequencies and decreasing amplitudes, the resulting FBM noise field exhibits scale-invariant, spatially coherent patterns that closely resemble real atmospheric fog. The generated FBM map serves as the simulated fog-intensity distribution for subsequent fog-mask construction and adversarial optimization.

\subsection{Fog Adversarial Example Optimization}
After constructing the simulated fog intensity map via multi-octave Perlin noise, we further optimize it to generate effective fog-based adversarial examples. Unlike conventional pixel-wise perturbations (e.g., FGSM \cite{goodfellow2014explaining}, BIM \cite{kurakin2018adversarial}, PGD \cite{madry2017towards}, MI-FGSM \cite{dong2018boosting}), FogFool restricts the perturbation space to a structured fog mask, thereby preserving physical plausibility while enabling adversarial manipulation.

\subsubsection{Fog-Based Image Formation}
Let $\mathbf{x} \in [0,1]^{H \times W \times C}$ denote the clean remote sensing image, and let $\mathbf{C} \in [0,1]^{H \times W \times C}$ represent the fog mask initialized using the FBM noise defined in Eq.~(\ref{eq:fbm_function}). 

To simulate atmospheric scattering, we construct a fog layer by blending the fog mask with a white color base:
\begin{equation}
\mathbf{F} = (1 - \lambda_w)\mathbf{C} + \lambda_w \mathbf{1},
\label{eq:fog_layer}
\end{equation}
where $\lambda_w \in [0,1]$ controls the fog whiteness intensity and $\mathbf{1}$ denotes an all-one tensor.

The adversarial image is then obtained via linear blending:
\begin{equation}
\mathbf{x}^{adv} = \lambda_b \mathbf{F} + (1 - \lambda_b)\mathbf{x},
\label{eq:image_blending}
\end{equation}
where $\lambda_b \in [0,1]$ controls the fog blending strength. The resulting image is clipped to ensure that pixel values fall within valid ranges.

\subsubsection{Gradient-Guided Fog Optimization}

The fog mask $\mathbf{C}$ is treated as the optimization variable. The objective is defined according to the adversarial attack setting.

For untargeted attacks, we maximize the classification loss:
\begin{equation}
\max_{\mathbf{C} \in [0,1]} 
\mathcal{L}\big(f(\mathbf{x}^{adv}), y\big);
\label{eq:fog_untargeted}
\end{equation}
while for targeted attacks, we minimize
\begin{equation}
\min_{\mathbf{C} \in [0,1]} 
\mathcal{L}\big(f(\mathbf{x}^{adv}), y_t\big).
\label{eq:fog_targeted}
\end{equation}
To solve this optimization problem, we adopt an iterative gradient-sign strategy inspired by the momentum-based iterative FGSM. At iteration $t$, the gradient with respect to the fog mask is computed as:
\begin{equation}
    \mathbf{g}^{(t)} = \nabla_{\mathbf{C}^{(t)}} 
    \mathcal{L}\big(f(\mathbf{x}^{adv}), y\big).
\end{equation}
The gradient is first normalized by its mean absolute value:
\begin{equation}
    \tilde{\mathbf{g}}^{(t)} = 
    \frac{\mathbf{g}^{(t)}}{\mathrm{mean}(|\mathbf{g}^{(t)}|)}.
\end{equation}
To stabilize optimization and improve transferability, a momentum term is incorporated:
\begin{equation}
\mathbf{m}^{(t)} = \mu \mathbf{m}^{(t-1)} + \tilde{\mathbf{g}}^{(t)},
\label{eq:momentum_update}
\end{equation}
where $\mu$ denotes the decay factor.

The fog mask is updated using the sign of the accumulated gradient for untargeted attacks (the sign is reversed for targeted attacks):
\begin{equation}
\mathbf{C}^{(t+1)} =
\Pi_{[0,1]} \Big(
\mathbf{C}^{(t)} + \alpha \cdot \mathrm{sign}(\mathbf{m}^{(t)})
\Big),
\label{eq:fog_update}
\end{equation}
where $\alpha$ is the step size and $\Pi_{[0,1]}$ denotes projection onto the valid intensity range.

\subsubsection{Fog Naturalness Regularization}
To preserve the spatial smoothness and natural appearance of fog, Gaussian filtering is applied to the updated fog mask after each iteration:
\begin{equation}
\mathbf{C}^{(t+1)} \leftarrow 
\mathcal{G}_{\sigma}\big(\mathbf{C}^{(t+1)}\big),
\label{eq:gaussian_smoothing}
\end{equation}
where $\mathcal{G}_{\sigma}(\cdot)$ denotes Gaussian smoothing with standard deviation $\sigma$. After $T$ iterations, the final adversarial example is obtained using Eqs.~(\ref{eq:fog_layer})–(\ref{eq:image_blending}) with the optimized fog mask.

\subsubsection{Algorithm Summary}
The overall fog adversarial optimization procedure is summarized in Algorithm~\ref{alg:fogfool}, which mainly consists of three core steps: constructing an initial fog intensity map via multi-octave Perlin noise, performing gradient-guided optimization to adjust the fog map for effective adversarial attacks while retaining natural fog characteristics, and iteratively updating and verifying until the termination condition is satisfied. The algorithm has a linear time complexity $O(H \times W \times C)$ (adaptive to input image size) due to fixed parameters $K$ and $T$, and it is highly practical with no complex preprocessing requirements, controllable parameters, and efficient generation of adversarial examples suitable for real-world remote sensing scenarios.

\begin{algorithm}[t]
\caption{Fog Adversarial Example Optimization}
\label{alg:fogfool}
\KwIn{Clean image $\mathbf{x}$, label $y$, classifier $f(\cdot)$}
\KwOut{Adversarial example $\mathbf{x}^{adv}$}

\nl Initialize fog mask $\mathbf{C}^{(0)}$ using multi-octave Perlin noise\;
\nl Initialize momentum $\mathbf{m}^{(0)} = 0$\;

\nl \For{$t = 0$ to $T-1$}{

    \nl Construct fog layer $\mathbf{F}$ using Eq.~(\ref{eq:fog_layer})\;
    
    \nl Generate adversarial image $\mathbf{x}^{adv}$ using Eq.~(\ref{eq:image_blending})\;
    
    \nl Compute loss $\mathcal{L}$ according to attack setting\;
    
    \nl Compute gradient $\mathbf{g}^{(t)} = \nabla_{\mathbf{C}^{(t)}} \mathcal{L}$\;
    
    \nl Normalize gradient\;
    
    \nl Update momentum using Eq.~(\ref{eq:momentum_update})\;
    
    \nl Update fog mask using Eq.~(\ref{eq:fog_update})\;
    
    \nl Apply Gaussian smoothing using Eq.~(\ref{eq:gaussian_smoothing})\;
}

\nl Construct final adversarial image using optimized fog mask\;

\nl \Return $\mathbf{x}^{adv}$\;
\end{algorithm}

\section{Experiments}
\label{sec:experiments}
To comprehensively evaluate the effectiveness of the proposed method, we conduct both untargeted and targeted adversarial attack experiments on two widely used remote sensing scene classification datasets: UC Merced Land Use (UCM)~\cite{yang2010bag} and NWPU-RESISC45 (NWPU)~\cite{cheng2017remote}.

\subsection{Datasets and Comparison Methods}
\subsubsection{Datasets}
\begin{table*}[htbp]
    \centering
    \caption{TEST ACCURACY (\%) OF DIFFERENT MODELS ON UCM AND NWPU DATASETS}
    \label{tab:models_acc}
    \renewcommand{\arraystretch}{1.2}
    \begin{tabular}{ccccccccc}
        \toprule
         Dataset & AlexNet & VGG16 & ResNet50 & ResNet101 & DenseNet121 & DenseNet201 & MobileNetV2 & EfficientNet-B0  \\
         \hline
         UCM & 88.33 & 93.33 & 91.67 & 95.24 & 94.05 & 93.10 & 90.71 & 92.62 \\
         NWPU & 90.89 & 93.10 & 95.52 & 95.99 & 95.73 & 96.17 & 95.03 & 95.99 \\
         \bottomrule
    \end{tabular}
\end{table*}

\textbf{UC Merced Land Use (UCM).}
The UCM dataset consists of 21 land-use scene categories, with 100 images per class. Each image has a spatial resolution of $256 \times 256$ pixels and a ground sampling distance of approximately 1 foot. The dataset was collected from the United States Geological Survey (USGS) National Map Urban Area Imagery collection, covering diverse urban regions.

\textbf{NWPU-RESISC45 (NWPU).}
The NWPU dataset is a large-scale benchmark for remote sensing scene classification, comprising 31,500 images across 45 scene categories, with 700 images per category. To ensure a fair evaluation of the proposed FogFool, the ``cloud'' category is excluded to avoid semantic overlap with the simulated fog perturbations. Consequently, 44 categories comprising 30,800 images are used in our experiments.

\subsubsection{Target Models}
For both datasets, we adopt a stratified split, allocating 80\% of the samples from each class to training and the remaining 20\% to testing. To comprehensively evaluate the effectiveness of FogFool across different architectures, we select eight widely adopted CNN models in remote sensing image classification, including AlexNet \cite{krizhevsky2012imagenet}, VGG16 \cite{simonyan2014very}, ResNet50 and ResNet101 \cite{he2016deep}, DenseNet121 and DenseNet201 \cite{huang2017densely}, MobileNetV2 \cite{sandler2018mobilenetv2}, and EfficientNet-B0 \cite{tan2019efficientnet}. The experimental platform is based on Ubuntu 22.04.5 LTS and the PyTorch framework \cite{paszke2019pytorch}, with Intel Xeon Gold 6342 24-Core Processor CPU and NVIDIA RTX A40 GPU. Each model is trained separately on the UCM and NWPU datasets based on pre-trained weights. The batch size is 32, and the learning rate is $10^{-4}$. The AdamW optimizer is employed with a weight decay of $10^{-4}$ for 50 training epochs. The input size of the images is 224 × 224, and data augmentation operations, e.g., random vertical or horizontal flips, are used during training. The test accuracy of each model on the two datasets is shown in Table \ref{tab:models_acc}.

\subsubsection{Comparison Methods}
To comprehensively evaluate the effectiveness of FogFool, we compare it with seven representative white-box attack methods, including FGSM \cite{goodfellow2014explaining}, BIM \cite{kurakin2018adversarial}, C\&W \cite{carlini2017towards}, PGD \cite{madry2017towards}, MI-FGSM \cite{dong2018boosting}, Jitter \cite{schwinn2023exploring}, and AutoAttack \cite{croce2020reliable}. 
To further assess transferability in the black-box setting, we adopt seven widely used transfer-based attacks, including MI-FGSM \cite{dong2018boosting}, DI-FGSM \cite{xie2019improving}, TI-FGSM \cite{dong2019evading}, NI-FGSM \cite{lin2019nesterov}, SI-NI-FGSM \cite{lin2019nesterov}, VMI-FGSM \cite{wang2021enhancing}, and VNI-FGSM \cite{wang2021enhancing}. 

\subsubsection{Metrics}
To quantitatively evaluate the attack performance, we adopt the Attack Success Rate (ASR) as the primary evaluation metric, defined as

\begin{equation}
\text{ASR} = \frac{n_{\text{adv}}}{n_{\text{total}} - n_{\text{mis}}}
\label{eq:ASR}
\end{equation}
where $n_{\text{total}}$ denotes the total number of test samples, and $n_{\text{mis}}$ represents the number of samples that are already misclassified by the model. These misclassified samples are excluded from evaluation because adversarial perturbations are unnecessary for them. $n_{\text{adv}}$ denotes the number of successfully generated adversarial examples that cause the target model to change its prediction.
For targeted attacks, ASR measures the proportion of adversarial examples that are successfully classified into the specified target class.
A higher ASR indicates stronger attack capability under the same perturbation constraint.

\subsection{Experimental Settings of the Proposed Method}
\begin{figure}[htbp]
    \centering
    \includegraphics[width=\linewidth]{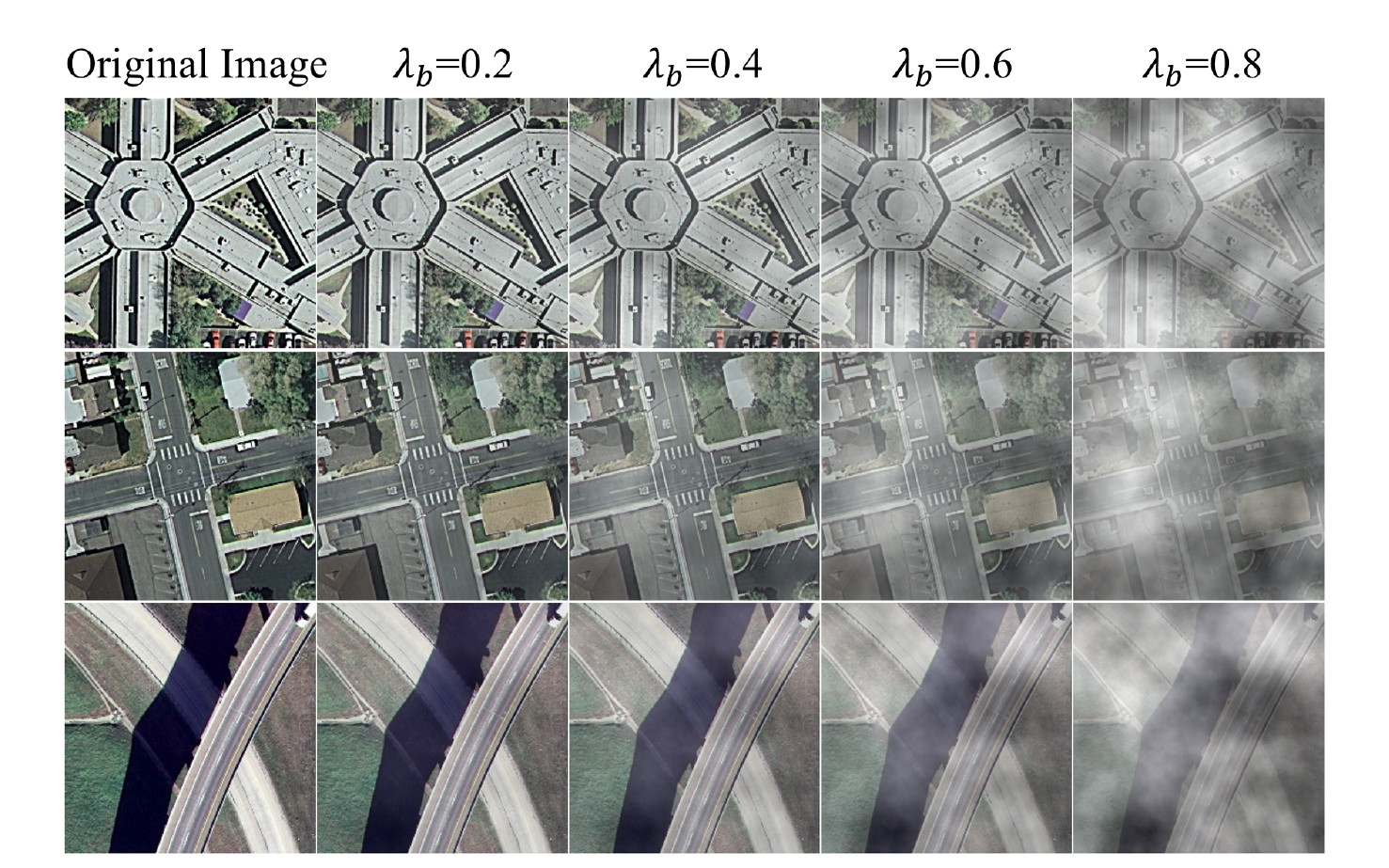}
    \caption{Illustration of fog-based adversarial examples under different fog blending coefficients $\lambda_b$. From left to right: clean image, $\lambda_b=0.2$, $\lambda_b=0.4$, $\lambda_b=0.6$, and $\lambda_b=0.8$. Increasing $\lambda_b$ intensifies the fog effect and increases the magnitude of the structured perturbation.}
    \label{fig:fog_blend}
\end{figure}

In this subsection, we present the implementation details and parameter configurations of FogFool. Unlike conventional adversarial attacks that constrain perturbations under a predefined $\ell_p$-norm bound, FogFool restricts perturbations to a physically plausible fog mask generated via multi-octave Perlin noise. No explicit perturbation budget (e.g., $\epsilon$ under $\ell_\infty$-norm) is imposed. Instead, the perturbation strength is implicitly controlled via fog-blending parameters and structured-mask optimization.

\subsubsection{Fog Initialization}
The initial fog mask is generated using FBM constructed from multi-octave Perlin noise. The number of octaves is set to $K=6$. The amplitude and frequency scaling factors follow exponential schedules defined as $\alpha_k = 2^{-k}$ and $\beta_k = 2^{k}$, respectively.

\subsubsection{Fog Formation Parameters}
The fog whiteness coefficient in Eq.~(\ref{eq:fog_layer}) is set to $\lambda_w = 0.2$, which controls the intensity of atmospheric scattering. The fog blending coefficient in Eq.~(\ref{eq:image_blending}) is set to $\lambda_b = 0.6$, balancing visibility preservation and attack strength. As demonstrated in Fig.~\ref{fig:fog_blend}, varying $\lambda_b$ produces progressively denser fog effects, demonstrating its direct influence on the perceptual strength of the generated adversarial examples.

\subsubsection{Optimization Settings}
\begin{table}[t]
    \centering
    \caption{Parameter Settings of the Proposed Fog-Based Attack}
    \label{tab:proposed_settings}
    \renewcommand{\arraystretch}{1.5} 
    \setlength{\tabcolsep}{5.0pt}  
    \begin{tabular}{cccccccccc}
    \hline
    \textbf{Parameter} & $K$ & $\alpha_k$ & $\beta_k$ & $\lambda_w$ & $\lambda_b$ & $T$ & $\alpha$ & $\mu$ & $\sigma$ \\
    \textbf{Value}     & 6   & $2^{-k}$   & $2^{k}$    & 0.2         & 0.6         & 20  & 1/255   & 1.0  & 0.7  \\
    \hline
    \end{tabular}
\end{table}

The fog mask is optimized for $T=20$ iterations using a gradient sign update strategy with momentum. The step size is set to $\alpha = 1/255$, and the momentum decay factor is $\mu = 1.0$. To preserve spatial smoothness and enhance perceptual realism, Gaussian smoothing with standard deviation $\sigma = 0.7$ is applied after each update. All parameter settings of the proposed method are summarized in Table~\ref{tab:proposed_settings}.

\subsection{Experimental Results and Analysis}
\subsubsection{Parameter Analysis}
\begin{figure}[h]
    \centering
    \includegraphics[width=\linewidth]{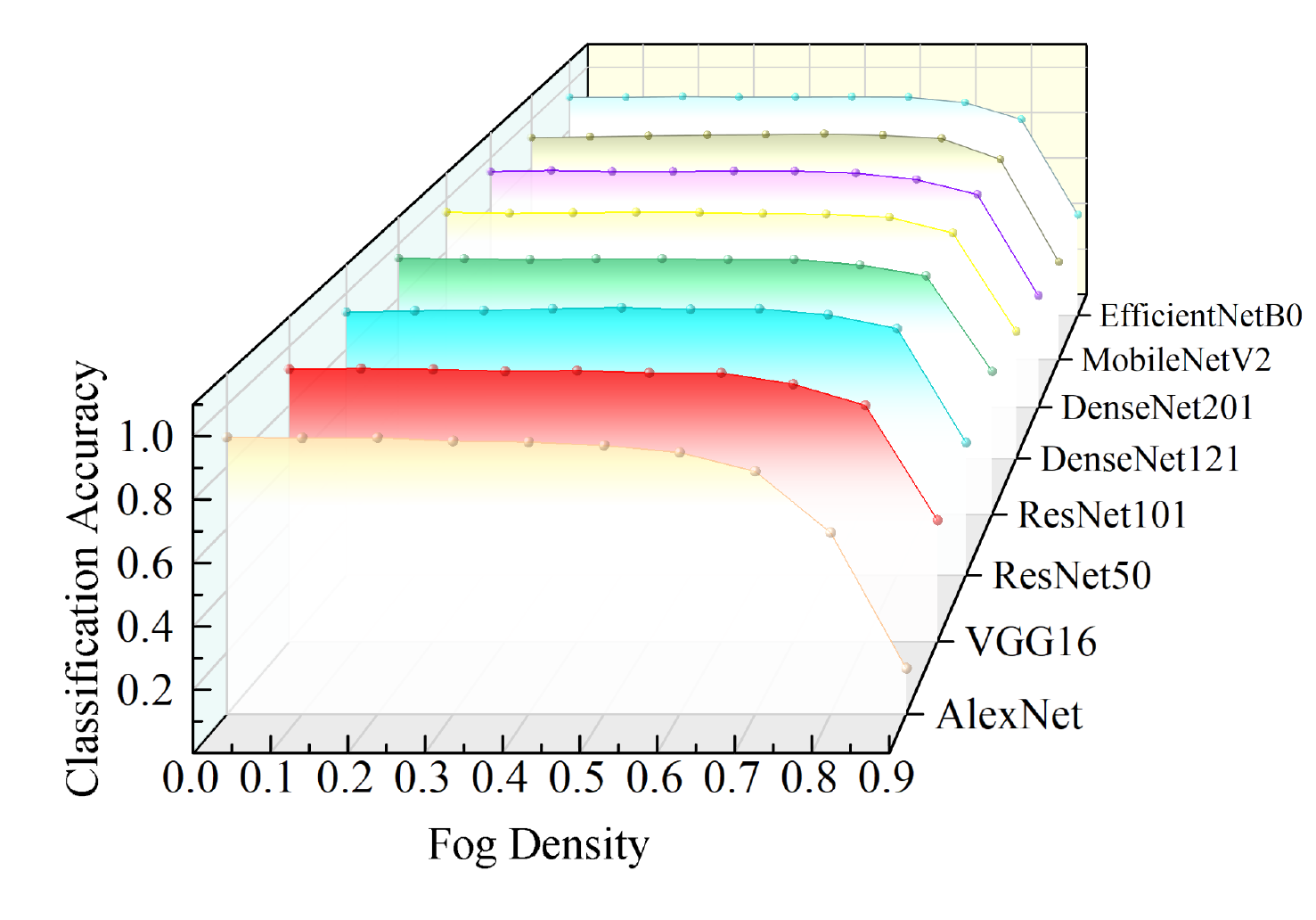}
    \caption{Classification accuracy of all evaluated models as a function of fog density $\lambda_b$, averaged across the UCM and NWPU datasets. Fog is applied without adversarial optimization to simulate natural atmospheric interference.}
    \label{fig:density_acc}
\end{figure}

To achieve a proper balance between attack effectiveness and visual plausibility, we conduct a systematic analysis of two key hyperparameters in the FogFool: the fog blending coefficient $\lambda_b$ and the number of optimization iterations $T$.

Fig.~\ref{fig:density_acc} presents the classification accuracy of all evaluated models on the UCM and NWPU datasets as $\lambda_b$ increases from 0 to 0.9, without applying adversarial optimization. When $\lambda_b = 0$, no fog blending is introduced and the models maintain their baseline accuracy, consistent with the results reported in Table~\ref{tab:models_acc}. As $\lambda_b$ increases, the fog intensity becomes progressively stronger, resulting in a gradual decline in classification accuracy. When $\lambda_b \le 0.6$, the accuracy degradation remains limited, with most models preserving over 90\% of their clean performance. Within this range, the fog perturbation remains visually plausible and does not substantially distort the semantic content of the remote sensing scenes. In contrast, when $\lambda_b > 0.6$, the accuracy drops significantly across models, indicating that excessive fog blending severely obscures discriminative features and compromises scene recognizability. Based on these observations, we adopt $\lambda_b = 0.6$ as the default setting. This choice provides a favorable trade-off, maintaining perceptual realism while preserving sufficient perturbation capacity for subsequent adversarial optimization.

\begin{figure}[t]
    \centering
    \includegraphics[width=\linewidth]{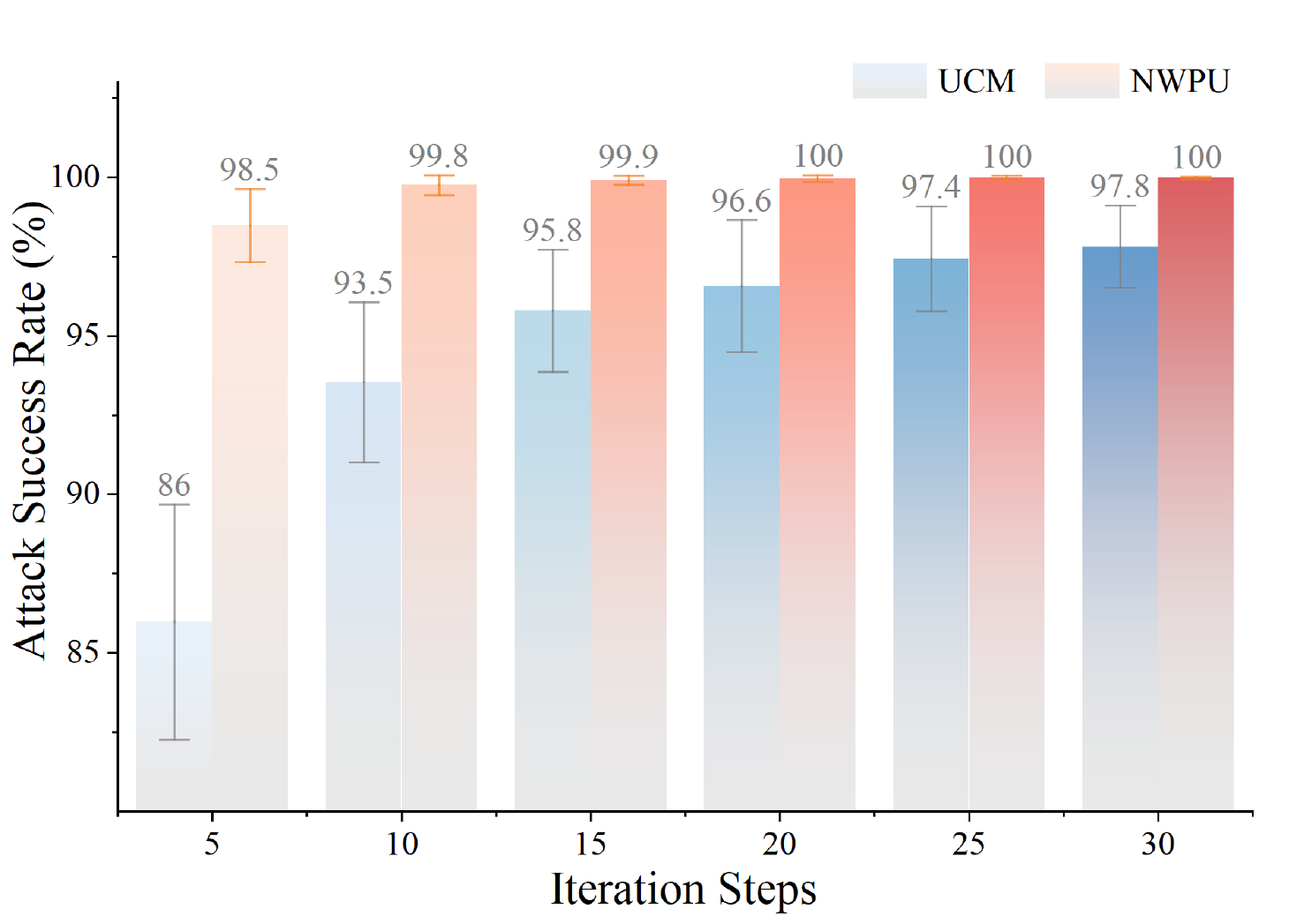}
    \caption{Attack success rate (ASR) versus the number of optimization iterations $T$ with $\lambda_b$ fixed at 0.6. Results are averaged across all evaluated models, with blue bars representing the UCM dataset and red bars representing the NWPU dataset. Error bars indicate the standard deviation of ASRs across different models.}
    \label{fig:steps_ASR}
\end{figure}

\begin{table*}[htbp]
\centering
\caption{Attack Success Rates (\%) of Different Methods on the UCM Dataset (Untargeted Attack) }
\label{tab:asr_ucm_untargeted}
\begin{tabular}{lcccccccccc}
\toprule
Attack method & AlexNet & VGG16 & ResNet50 & ResNet101 & DenseNet121 & DenseNet201 & MobileNetV2 & EfficientNet-B0 & Average \\
\midrule
FGSM & 90.84 & 74.49 & 43.64 & 56.00 & 57.72 & 60.61 & 57.48 & 82.26 & 65.38 \\
BIM & 100.00 & 98.98 & 100.00 & 100.00 & 99.75 & 100.00 & 99.21 & 100.00 & 99.74 \\
CW & 79.78 & 90.05 & 95.32 & 97.00 & 95.95 & 96.93 & 94.49 & 96.92 & 93.31 \\
PGD & 100.00 & 98.72 & 100.00 & 100.00 & 100.00 & 100.00 & 99.48 & 100.00 & 99.77 \\
MIFGSM & 100.00 & 98.47 & 100.00 & 99.50 & 99.49 & 100.00 & 99.74 & 100.00 & 99.65 \\
Jitter & 90.84 & 82.65 & 95.32 & 94.75 & 95.95 & 94.88 & 97.90 & 99.49 & 93.97 \\
AutoAttack & 100.00 & 100.00 & 100.00 & 100.00 & 100.00 & 100.00 & 99.74 & 98.71 & \textbf{99.81} \\
FogFool & 98.11 & 93.62 & 96.62 & 96.25 & 96.46 & 98.98 & 97.90 & 96.40 & 96.79 \\
\bottomrule
\end{tabular}
\end{table*}

\begin{table*}[htbp]
\centering
\caption{Attack Success Rates (\%) of Different Methods on the UCM Dataset (Targeted Attack)}
\label{tab:asr_ucm_targeted}
\begin{tabular}{lcccccccccc}
\toprule
Attack method & AlexNet & VGG16 & ResNet50 & ResNet101 & DenseNet121 & DenseNet201 & MobileNetV2 & EfficientNet-B0 & Average \\
\midrule
BIM & 97.57 & 95.66 & 98.70 & 99.00 & 99.49 & 99.49 & 98.43 & 99.23 & 98.45 \\
PGD & 96.23 & 94.64 & 98.18 & 98.25 & 99.49 & 99.74 & 98.69 & 98.71 & 97.99 \\
MIFGSM & 97.84 & 97.45 & 100.00 & 99.50 & 99.49 & 99.74 & 98.43 & 100.00 & \textbf{99.06} \\
FogFool & 96.23 & 92.35 & 98.18 & 99.25 & 99.49 & 100.00 & 97.38 & 98.46 & 97.67 \\
\bottomrule
\end{tabular}
\end{table*}

\begin{table*}[htbp]
\centering
\caption{Attack Success Rates (\%) of Different Methods on the NWPU Dataset (Untargeted Attack)}
\label{tab:asr_nwpu_untargeted}
\begin{tabular}{lcccccccccc}
\toprule
Attack method & AlexNet & VGG16 & ResNet50 & ResNet101 & DenseNet121 & DenseNet201 & MobileNetV2 & EfficientNet-B0 & Average \\
\midrule
FGSM & 91.12 & 87.20 & 73.28 & 80.33 & 84.25 & 83.79 & 82.30 & 82.21 & 83.06 \\
BIM & 99.95 & 99.69 & 100.00 & 100.00 & 100.00 & 100.00 & 100.00 & 100.00 & 99.95 \\
CW & 84.10 & 88.88 & 86.15 & 83.58 & 93.35 & 90.61 & 96.24 & 96.36 & 89.91 \\
PGD & 99.96 & 99.81 & 100.00 & 100.00 & 100.00 & 100.00 & 100.00 & 100.00 & 99.97 \\
MIFGSM & 99.95 & 99.69 & 100.00 & 100.00 & 100.00 & 100.00 & 100.00 & 100.00 & 99.95 \\
Jitter & 99.96 & 99.84 & 98.88 & 99.04 & 98.34 & 97.79 & 99.71 & 99.90 & 99.18 \\
AutoAttack & 100.00 & 100.00 & 100.00 & 100.00 & 100.00 & 100.00 & 100.00 & 100.00 & \textbf{100.00} \\
FogFool & 99.96 & 99.72 & 100.00 & 100.00 & 100.00 & 100.00 & 100.00 & 100.00 & 99.96 \\
\bottomrule
\end{tabular}
\end{table*}

\begin{table*}[htbp]
\centering
\caption{Attack Success Rates (\%) of Different Methods on the NWPU Dataset (Targeted Attack)}
\label{tab:asr_nwpu_targeted}
\begin{tabular}{lcccccccccc}
\toprule
Attack method & AlexNet & VGG16 & ResNet50 & ResNet101 & DenseNet121 & DenseNet201 & MobileNetV2 & EfficientNet-B0 & Average \\
\midrule
BIM & 99.48 & 99.18 & 99.88 & 99.81 & 100.00 & 100.00 & 100.00 & 99.97 & 99.79 \\
PGD & 99.46 & 99.13 & 99.90 & 99.85 & 100.00 & 100.00 & 100.00 & 99.98 & 99.79 \\
MIFGSM & 99.93 & 99.84 & 100.00 & 100.00 & 100.00 & 100.00 & 100.00 & 100.00 & \textbf{99.97} \\
FogFool & 99.96 & 99.51 & 100.00 & 100.00 & 100.00 & 100.00 & 100.00 & 100.00 & 99.93 \\
\bottomrule
\end{tabular}
\end{table*}
Fig.~\ref{fig:steps_ASR} illustrates the ASR of FogFool as a function of the number of optimization iterations $T$, with $\lambda_b$ fixed at 0.6. On both datasets, the ASR increases rapidly during the initial iterations, demonstrating that the gradient-guided optimization effectively refines the structured fog mask to induce misclassification. As $T$ increases, the ASR gradually saturates, and the performance gain becomes marginal beyond approximately 20 iterations. This convergence pattern is consistently observed across different model architectures, indicating that the optimization process stabilizes within a relatively small number of update steps. Considering both attack performance and computational efficiency, we set the number of iterations to $T = 20$ in subsequent experiments. This configuration achieves near-saturated ASR while keeping the computational overhead moderate.

\subsubsection{Quantitative Analysis}
\begin{figure*}[!t]
    \centering
    \includegraphics[width=\linewidth]{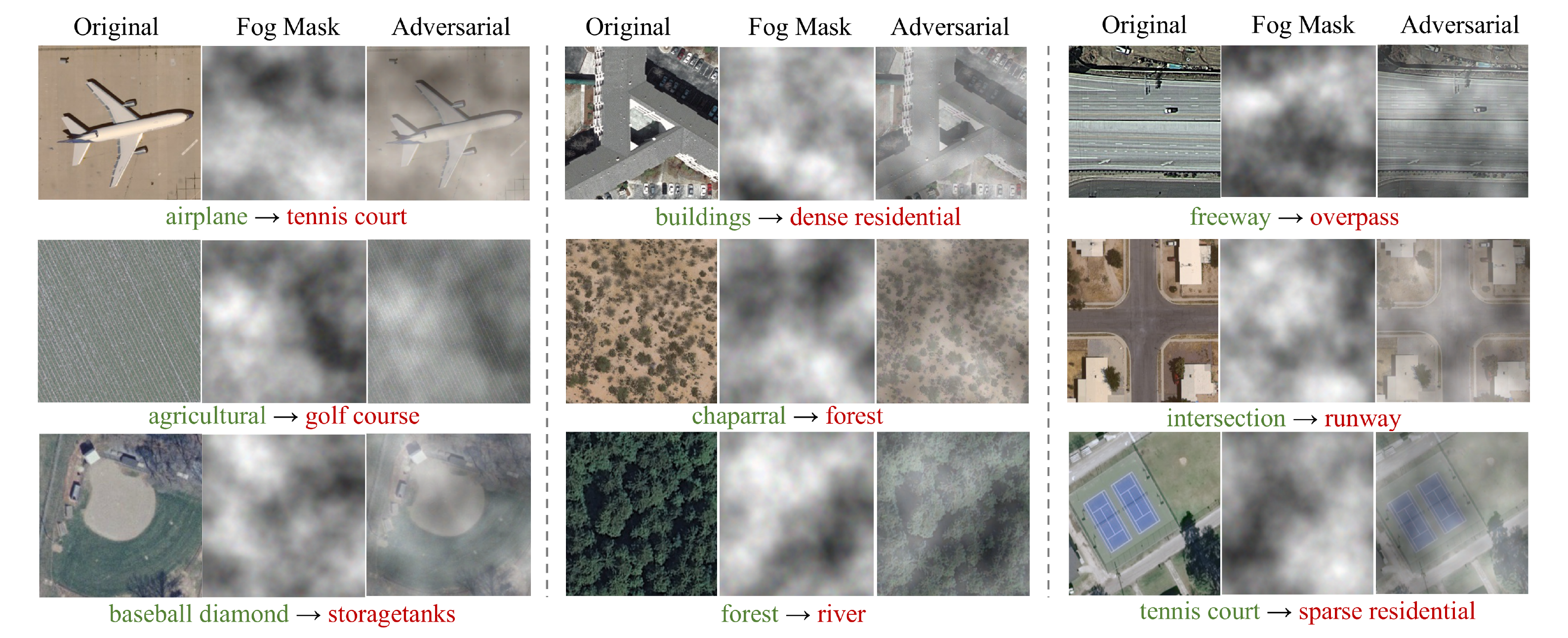}
    \caption{Adversarial examples generated by the proposed method on the UCM dataset. Each group of images, from left to right, the original images, the Perlin noise fog masks, and the corresponding adversarial examples. Green labels indicate the correct categories and red labels indicate the labels predicted by the target models.}
    \label{fig:UCM_ADVs}
\end{figure*}
\begin{figure*}[!t]
    \centering
    \includegraphics[width=\linewidth]{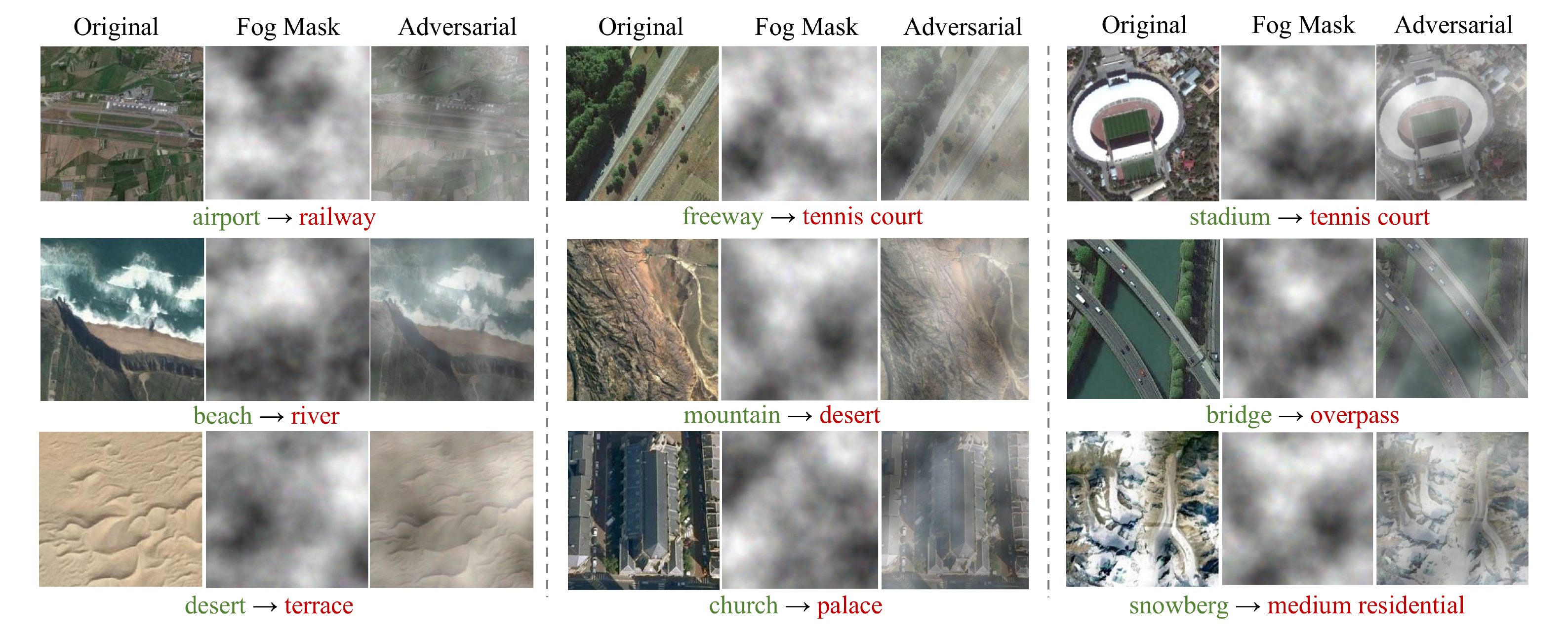} 
    \caption{Adversarial examples generated by the proposed method on the NWPU dataset. Each group of images, from left to right, the original images, the Perlin noise fog masks, and the corresponding adversarial examples. Green labels indicate the correct categories and red labels indicate the labels predicted by the target models.}
    \label{fig:NWPU_ADVs}
\end{figure*}

We proceed to comprehensively evaluate the attack performance of the proposed FogFool method in both untargeted and targeted settings. The quantitative results are summarized in Tables~\ref{tab:asr_ucm_untargeted}–\ref{tab:asr_nwpu_targeted}, and representative adversarial examples are illustrated in Figs.~\ref{fig:UCM_ADVs} and \ref{fig:NWPU_ADVs}. As shown in Table \ref{tab:asr_ucm_untargeted}, FogFool achieves an average ASR of 96.79\% on the UCM dataset, maintaining consistent performance across all 8 models (93.62\%–98.98\%). On the NWPU dataset (Table \ref{tab:asr_nwpu_untargeted}), FogFool reaches an average ASR of 99.96\%, comparable to the state-of-the-art methods (PGD: 99.97\%, AutoAttack: 100.00\%), with 100\% ASR on 6 models, demonstrating effective untargeted attack capability.

Tables \ref{tab:asr_ucm_targeted} and \ref{tab:asr_nwpu_targeted} present targeted attack results. On the UCM dataset, FogFool achieves an average ASR of 97.67\%, with a maximum of 100.00\% on DenseNet201. On the NWPU dataset, it reaches 99.93\% average ASR, only 0.04\% lower than the top-performing MI-FGSM, confirming qualified targeted attack performance.

Figs. \ref{fig:UCM_ADVs} and \ref{fig:NWPU_ADVs} show that FogFool generates visually natural adversarial examples. The Perlin noise-based fog masks are spatially coherent and realistic, and the adversarial samples maintain physical plausibility without obvious artifacts, while effectively misleading target models.

Overall, the experimental results demonstrate that FogFool achieves competitive ASR under both untargeted and targeted settings across two benchmark remote sensing datasets. Importantly, these results are obtained without relying on explicit $\ell_p$-norm perturbation constraints. Instead, the attack operates within a physically interpretable fog generation framework, offering a favorable balance between attack strength and visual realism. This highlights the potential vulnerability of remote sensing models to semantically meaningful and physically plausible adversarial manipulations.

\subsubsection{Attack Selectivity}
\begin{figure*}[!t]
    \centering
    \includegraphics[width=\linewidth]{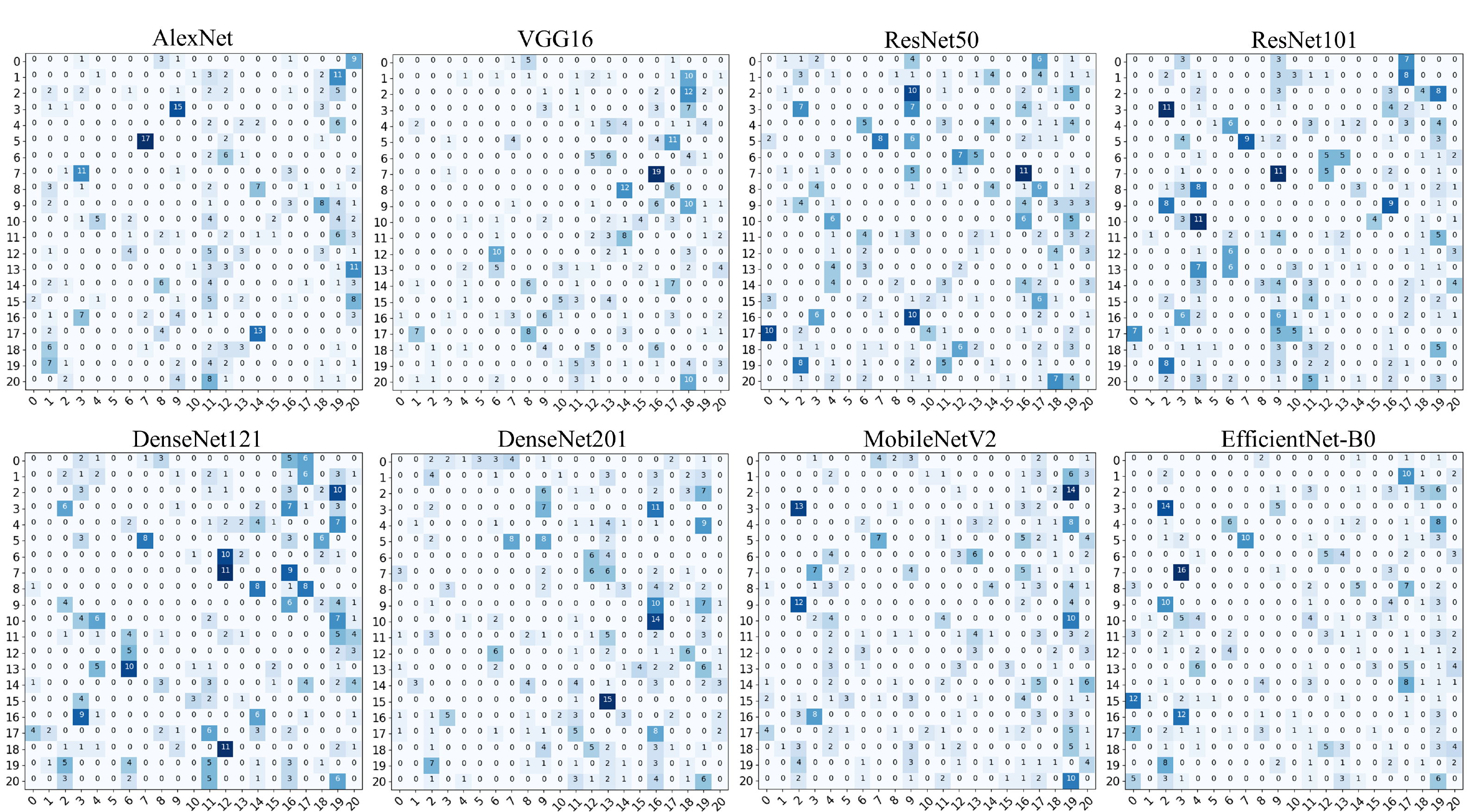}
    \caption{Confusion matrices of fog adversarial examples on the UCM dataset for different target models. Rows represent ground-truth labels, and columns represent predicted labels. The label numbers from 0 to 20 represent the categories as follows: agricultural, airplane, baseball diamond, beach, buildings, chaparral, dense residential, forest, freeway, golf course, harbor, intersection, medium residential, mobile home park, overpass, parking lot, river, runway, sparse residential, storage tanks, and tennis court.}
    \label{fig:confusion_matrix}
\end{figure*}

Chen et al.~\cite{chen2020attack} introduced the concept of \emph{attack selectivity} in remote sensing image classification, showing that adversarial misclassifications are not uniformly distributed across categories but instead concentrate on specific target classes. To further investigate this phenomenon, Fig.~\ref{fig:confusion_matrix} presents the confusion matrices of fog adversarial examples on the UCM dataset. 

Notably, the selective misclassification behavior varies across model architectures. For instance, ResNet50 exhibits a strong tendency to misclassify adversarial examples into category 9 (Golf Course), whereas DenseNet201 shows a clear bias toward category 16 (River). This suggests that attack selectivity is closely related to the geometric properties of the decision boundaries learned by different neural network architectures. Under fog-induced perturbations, feature representations may be systematically shifted toward particular regions of the classification space.

Furthermore, intrinsic semantic similarity between categories also plays a crucial role. Regardless of the target model, samples from category 5 (Chaparral) are predominantly misclassified as category 7 (Forest). Given the high visual and semantic similarity between these land-cover types, their corresponding feature embeddings are likely to be closely distributed in high-dimensional representation space. Consequently, small structured perturbations can more easily push such samples across decision boundaries into semantically adjacent categories.

\subsubsection{Transfer Attack}
\begin{table*}[htbp]
\centering
\caption{Ensemble-based Transfer Attack Success Rates (\%) on the UCM Dataset}
\label{tab:transfer_ucm}
\begin{tabular}{l|ccc|ccccc|c}
\toprule
Attack method & ResNet50 & DenseNet121 & MobileNetV2 & AlexNet & VGG16 & ResNet101 & DenseNet201 & EfficientNet-B0 & Average \\
\midrule
MI-FGSM & 100.00 & 95.70 & 98.16 & 12.40 & 24.49 & 39.50 & 36.06 & 62.98 & 35.08 \\
DI-FGSM & 96.88 & 94.18 & 96.85 & 15.09 & 35.20 & 60.75 & 54.99 & 67.87 & 46.78 \\
TI-FGSM & 98.18 & 94.68 & 95.54 & 19.41 & 40.56 & 61.75 & 62.40 & 68.89 & 50.60 \\
NI-FGSM & 98.44 & 94.94 & 98.16 & 11.86 & 23.21 & 38.00 & 34.27 & 64.78 & 34.42 \\
SI-NI-FGSM & 97.14 & 95.70 & 97.38 & 13.21 & 28.32 & 47.25 & 39.64 & 64.78 & 38.64 \\
VMI-FGSM & 100.00 & 95.44 & 98.69 & 15.90 & 34.95 & 59.75 & 56.01 & 69.67 & 47.25 \\
VNI-FGSM & 100.00 & 96.20 & 98.16 & 15.63 & 35.97 & 62.75 & 57.29 & 70.95 & 48.51 \\
FogFool & 92.99 & 91.65 & 91.60 & 44.20 & 41.84 & 56.75 & 56.52 & 58.61 & \textbf{51.58} \\
\bottomrule
\end{tabular}
\end{table*}

\begin{table*}[htbp]
\centering
\caption{Ensemble-based Transfer Attack Success Rates (\%) on the NWPU Dataset}
\label{tab:transfer_nwpu}
\begin{tabular}{l|ccc|ccccc|c}
\toprule
Attack method & ResNet50 & DenseNet121 & MobileNetV2 & AlexNet & VGG16 & ResNet101 & DenseNet201 & EfficientNet-B0 & Average \\
\midrule
MI-FGSM & 99.97 & 99.86 & 100.00 & 26.90 & 49.78 & 72.87 & 73.21 & 62.74 & 57.10 \\
DI-FGSM & 99.93 & 99.61 & 100.00 & 37.61 & 69.08 & 86.34 & 85.16 & 81.04 & 71.84 \\
TI-FGSM & 99.86 & 99.61 & 100.00 & 48.44 & 73.67 & 86.96 & 87.59 & 84.63 & 76.25 \\
NI-FGSM & 99.93 & 99.83 & 100.00 & 25.63 & 48.20 & 69.64 & 67.59 & 59.83 & 54.17 \\
SI-NI-FGSM & 98.88 & 99.19 & 100.00 & 29.34 & 52.17 & 70.89 & 67.00 & 64.06 & 56.69 \\
VMI-FGSM & 99.97 & 99.66 & 100.00 & 37.19 & 66.70 & 84.56 & 83.19 & 79.00 & 70.12 \\
VNI-FGSM & 99.98 & 99.76 & 100.00 & 38.13 & 70.78 & 87.13 & 85.25 & 80.65 & 72.38 \\
FogFool & 99.63 & 99.71 & 99.93 & 74.64 & 76.13 & 89.13 & 91.31 & 87.49 & \textbf{83.74} \\
\bottomrule
\end{tabular}
\end{table*}

\begin{table*}[htbp]
\centering
\caption{Average Linear CKA Values of Feature Discrepancy Between Surrogate Models and Target Models}
\label{tab:cka}
\begin{tabular}{llcccccc}
\toprule
Dataset & Attack Method & AlexNet & VGG16 & ResNet101 & DenseNet201 & EfficientNet-B0 & Average \\
\midrule
\multirow{8}{*}{\makecell[l]{UCM}}
& MI-FGSM    & 0.1243 & 0.0298 & 0.3865 & 0.4335 & 0.3179 & 0.2584 \\
& DI-FGSM    & 0.1812 & 0.0722 & 0.5886 & 0.5756 & 0.4553 & 0.3746 \\
& TI-FGSM    & 0.1900 & 0.0781 & 0.5857 & 0.5727 & 0.4800 & 0.3813 \\
& NI-FGSM    & 0.1372 & 0.0296 & 0.5617 & 0.5191 & 0.3244 & 0.3144 \\
& SI-NI-FGSM & 0.1738 & 0.0365 & 0.5831 & 0.6179 & 0.4541 & 0.3731 \\
& VMI-FGSM   & 0.1449 & 0.0443 & 0.4682 & 0.5000 & 0.4356 & 0.3186 \\
& VNI-FGSM   & 0.1724 & 0.0740 & 0.6229 & 0.5983 & 0.5034 & 0.3942 \\
& FogFool    & 0.2245 & 0.0458 & 0.6070 & 0.5657 & 0.5291 & \textbf{0.3944} \\
\midrule
\multirow{8}{*}{\makecell[l]{NWPU}}
& MI-FGSM    & 0.1791 & 0.0831 & 0.4363 & 0.4308 & 0.2836 & 0.2826 \\
& DI-FGSM    & 0.2080 & 0.1065 & 0.5721 & 0.5340 & 0.3610 & 0.3563 \\
& TI-FGSM    & 0.2168 & 0.1110 & 0.5768 & 0.5882 & 0.4189 & 0.3823 \\
& NI-FGSM    & 0.1951 & 0.1130 & 0.4761 & 0.4445 & 0.3043 & 0.3066 \\
& SI-NI-FGSM & 0.2255 & 0.1727 & 0.5509 & 0.4952 & 0.3758 & 0.3640 \\
& VMI-FGSM   & 0.2083 & 0.1210 & 0.5420 & 0.5383 & 0.3855 & 0.3590 \\
& VNI-FGSM   & 0.2235 & 0.1291 & 0.5634 & 0.5538 & 0.4040 & 0.3748 \\
& FogFool    & 0.2102 & 0.1053 & 0.6527 & 0.6498 & 0.6273 & \textbf{0.4490} \\
\bottomrule
\end{tabular}
\end{table*}

To evaluate the cross-model transferability of the FogFool, we conduct ensemble-based transfer attack experiments. Specifically, three models (i.e., ResNet50, DenseNet121, and MobileNet\_V2) are selected as surrogate models to generate adversarial examples in a white-box setting. The generated samples are then directly applied to attack five unseen target models in the black-box setting. The transfer attack success rates (TASRs) on the UCM and NWPU datasets are reported in Tables~\ref{tab:transfer_ucm} and \ref{tab:transfer_nwpu}, respectively.

As shown in Table~\ref{tab:transfer_ucm}, FogFool achieves an average TASR of 51.58\% on the UCM dataset, outperforming all compared gradient-based ensemble methods. Although the ASRs on the three surrogate models are slightly lower than those of iterative gradient-based attacks, FogFool exhibits substantially improved transferability on the five unseen target models. In particular, the TASRs on AlexNet and VGG16 are noticeably higher than those of the other methods. This suggests that the proposed fog-based perturbations using FogFool do not overly adapt to the surrogate models' gradients. Instead, they induce more globally effective feature shifts that generalize across heterogeneous architectures, thereby enhancing cross-model transferability.

On the larger and more complex NWPU dataset (Table~\ref{tab:transfer_nwpu}), the superiority of the proposed FogFool becomes even more evident. FogFool achieves an average TASR of 83.74\%, significantly exceeding all baseline attacks. Moreover, the TASRs on all five unseen target models consistently exceed 74\%, reaching up to 91.31\% on DenseNet201. Compared with conventional gradient-based ensemble attacks, which tend to rely heavily on surrogate-specific gradient alignment, FogFool generates structured, low-frequency fog perturbations that resemble realistic atmospheric effects. Such perturbations are more likely to disrupt shared and robust features learned across different architectures, leading to stronger transferability.

To further reveal the intrinsic mechanism underlying the superior transferability of FogFool, we adopt linear Centered Kernel Alignment (CKA) to quantify the consistency of intermediate feature deviations across models. As reported in Tables~\ref{tab:cka} , We first extract deep backbone features of both clean images and adversarial examples from the surrogate and target networks, and compute the feature difference between adversarial and clean representations to characterize layer-wise perturbation offsets. Then, linear CKA is conducted on these intermediate deviation maps between the surrogate models and black-box targets, with averaged CKA scores adopted to evaluate the similarity of cross-model feature shifting directions.

Experimental results on both UCM and NWPU datasets consistently demonstrate that FogFool achieves the highest average CKA value among all attack methods. This indicates that the fog-based adversarial examples generated by FogFool can induce much more similar intermediate feature discrepancy directions across the surrogate models and diverse target models, compared with traditional pixel-wise gradient-based attacks. The structured, low-frequency and physically plausible fog perturbations optimize the universal feature space shared by different network architectures, rather than overfitting to the specific gradient characteristics of the surrogate models. This consistent feature deviation pattern across models explains why FogFool possesses outstanding cross-model transferability in black-box remote sensing image classification scenarios.

\subsubsection{Defense Against Perlin Noise Fog}
\begin{figure*}
    \centering
    \includegraphics[width=\linewidth]{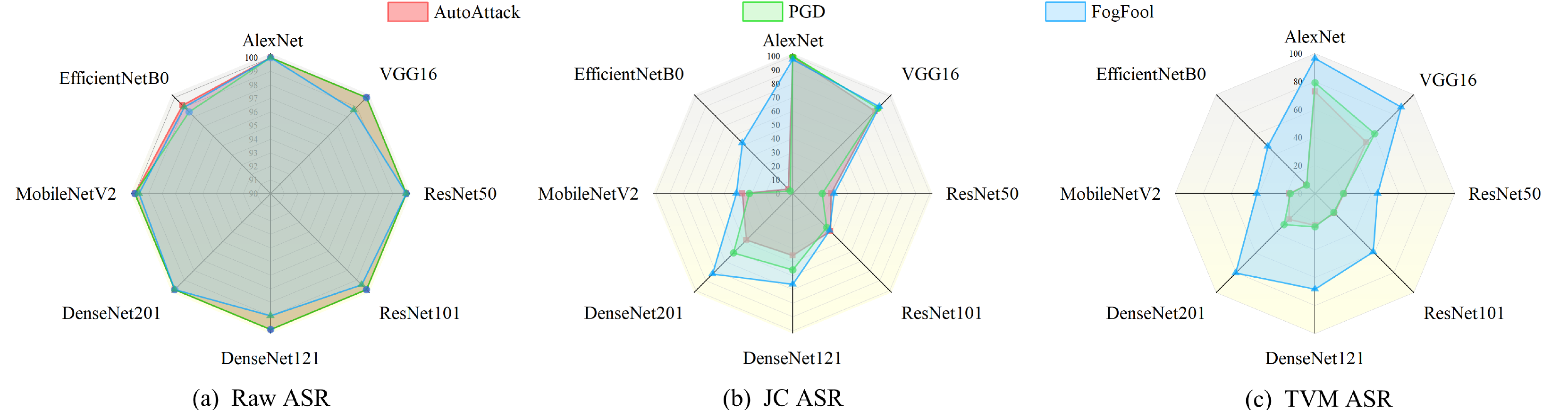}
    \caption{Radar charts comparing the attack success rates (ASRs) of AutoAttack, PGD, and the proposed FogFool method across eight models. (a) Raw ASR without defense, (b) ASR under JPEG compression (JC) defense, and (c) ASR under total variation minimization (TVM) defense. Results are evaluated on the UCM dataset.}
    \label{fig:robust}
\end{figure*}

To further evaluate the robustness of the proposed FogFool under adversarial defense scenarios, we compare FogFool with two strong white-box baseline attacks, AutoAttack and PGD. First, adversarial examples are generated using the three attack methods to obtain the original ASRs. Subsequently, two commonly used defense preprocessing techniques—JPEG compression (JC)~\cite{dziugaite2016study} and total variation minimization (TVM)~\cite{guo2017countering}—are applied to mitigate adversarial perturbations. The ASRs are then re-evaluated after defense processing. For TVM, the reconstruction method is set to Bregman iteration~\cite{goldstein2009split}, the pixel random drop rate is 0.5, and the TV regularization weight is 0.03. For JPEG compression, the image quality factor is set to 50. The experimental results are illustrated in Fig.~\ref{fig:robust}.

As shown in Fig.~\ref{fig:robust}, FogFool exhibits stronger resilience against both JC and TVM defenses compared with AutoAttack and PGD. Under JPEG compression, the ASRs of AutoAttack and PGD decrease substantially across most architectures, whereas FogFool maintains relatively higher ASRs values. For instance, on ResNet50, the ASRs of AutoAttack and PGD decrease to 28.1\% and 21.6\%, respectively, while FogFool retains 30.1\%. The performance gap further widens under TVM defense. On DenseNet121, the ASRs of AutoAttack and PGD decrease to 22.8\% and 23.8\%, respectively, whereas FogFool achieves 68.3\%, indicating significantly stronger robustness against smoothing-based defenses.

This improved resilience can be attributed to the spatially coherent and low-frequency characteristics of the Perlin noise-based fog perturbations. Unlike pixel-wise gradient perturbations, which often contain high-frequency components that are easily suppressed by compression or denoising operations, FogFool resemble natural atmospheric scattering patterns. Such structured perturbations are less sensitive to frequency-domain filtering and smoothing processes, thereby maintaining higher attack effectiveness after defense preprocessing. These results demonstrate that FogFool not only achieves competitive attack performance but also preserves stronger robustness under common adversarial defense strategies.

\subsubsection{Why Perlin Noise Fog Works}
\begin{figure}
    \centering
    \includegraphics[width=\linewidth]{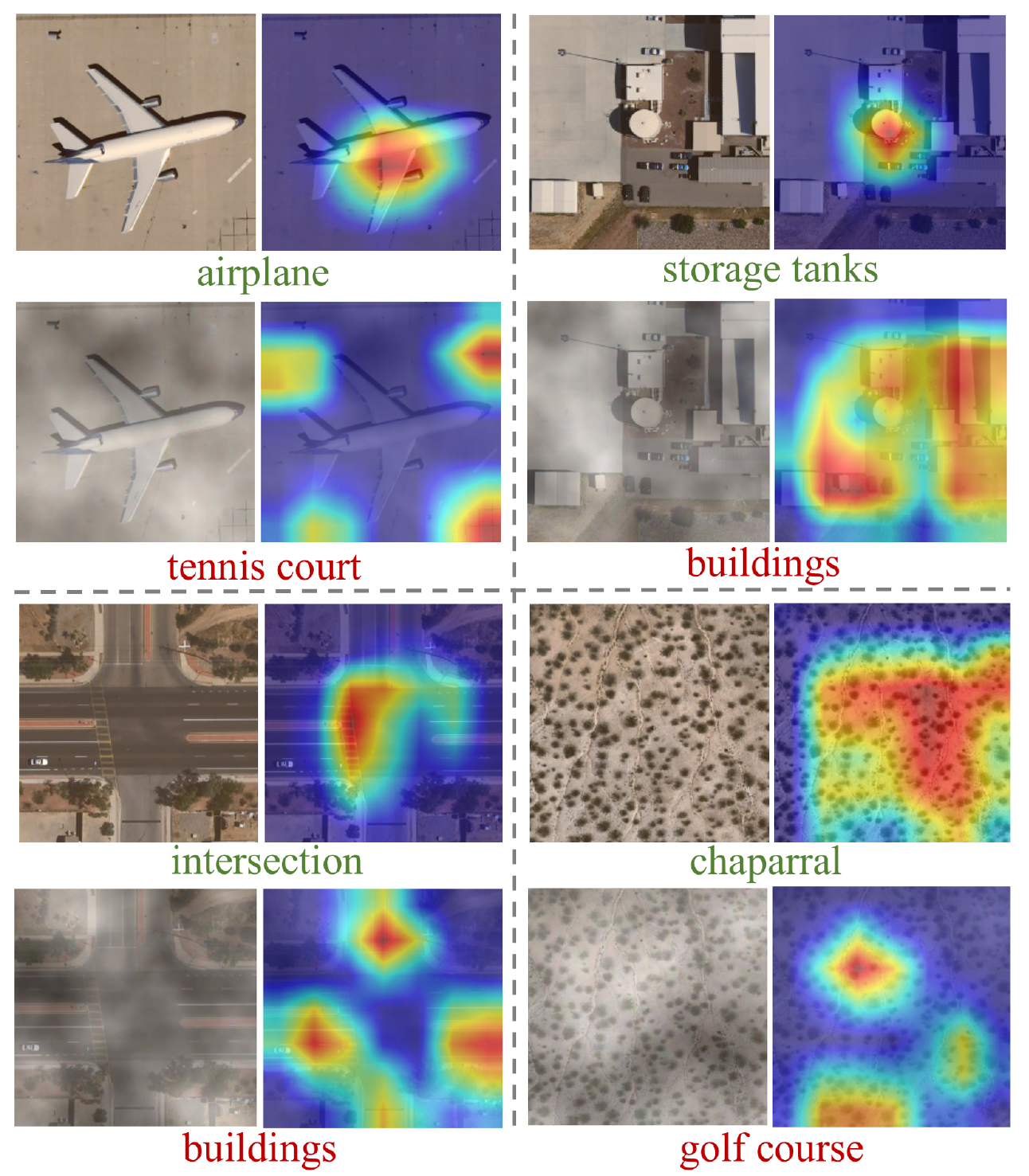}
    \caption{Grad-CAM visualizations of ResNet50 on the UCM dataset. From (top to bottom), original images with their CAMs, and adversarial examples with their CAMs. The green and red labels indicate the true categories and predicted results by the target models.}
    \label{fig:CAMs}
\end{figure}

To investigate the underlying mechanism of FogFool, we employ Grad-CAM~\cite{selvaraju2017grad} to visualize the class activation maps of ResNet50 before and after applying Perlin noise fog perturbations. Grad-CAM highlights the discriminative regions that contribute most significantly to the model’s prediction.

As shown in Fig.~\ref{fig:CAMs}, the introduction of Perlin noise fog leads to a substantial redistribution of attention in the feature space. In the original images, the network focuses primarily on semantically meaningful objects, such as the airplane in the leftmost example. However, after adding structured fog perturbations, the high-activation regions shift toward surrounding background areas, while the object-related activations are weakened.

Importantly, the global structural integrity of the objects remains largely preserved in the adversarial examples, indicating that the perturbation does not rely on visually destructive distortions. Instead, the Perlin noise fog alters intermediate feature representations and attention allocation, thereby disrupting the discriminative cues used by the classifier. This structured redistribution of activation regions explains why the proposed fog-based perturbations can effectively mislead the neural network while maintaining perceptual plausibility.

\section{Conclusion}
\label{sec:conclusion}

In this paper, we proposed FogFool, a physically plausible adversarial attack framework for remote sensing image classification. Unlike conventional pixel-wise perturbation methods constrained by $\ell_p$-norms, FogFool restricts the perturbation space to structured atmospheric fog patterns generated via multi-octave Perlin noise and fractional brownian motion. By integrating procedural fog simulation with gradient-guided optimization, FogFool is capable of producing both untargeted and targeted adversarial examples while maintaining strong spatial coherence and visual realism.
Extensive experiments conducted on two benchmark remote sensing datasets demonstrate that FogFool achieves competitive or superior ASRs compared with state-of-the-art white-box and transfer-based attack methods. FogFool maintains consistently high ASRs across diverse network architectures, including lightweight and deep convolutional models. Furthermore, parameter analyses confirm that the fog blending coefficient and optimization iterations can effectively balance visual plausibility and attack effectiveness. The generated adversarial examples exhibit natural atmospheric characteristics, avoiding the artificial high-frequency artifacts commonly observed in traditional attacks.  As a potential future direction, we are looking forward to extending our method to improve the performance of various applications, such as large language
models~\cite{lin2024splitlora,fang2026hfedmoe,lin2024split,qu2025mobile,fang2024automated} and distributed learning
system~\cite{lin2025hierarchical,zhang2024satfed,hong2026conflict,lin2025hasfl,hu2024accelerating,fang2026aggregation,lyu2023optimal,lin2024adaptsfl}.

\bibliographystyle{IEEEtran}
\bibliography{ref}
\end{document}